# Automated Invoice Data Extraction: Using LLM and OCR

Khushi Khanchandani[1], Advait Thakur*[2], Akshita Shetty[2], Chaitravi Reddy[2], and Ritisa Behera[2]

[1] Assistant Professor, Department of Information Technology, K.J. Somaiya School of Engineering (formerly K. J. Somaiya College of Engineering), Somaiya Vidyavihar University, Mumbai, India

[2] Undergraduate Student, Department of Information Technology, K.J. Somaiya School of Engineering (formerly K. J. Somaiya College of Engineering), Somaiya Vidyavihar University, Mumbai, India

* Corresponding Author's Email: advait.thakur@somaiya.edu

## Abstract

Conventional Optical Character Recognition (OCR) systems are challenged by variant invoice layouts, handwritten text, and low-quality scans, which are often caused by strong template dependencies that restrict their flexibility across different document structures and layouts. Newer solutions utilize advanced deep learning models such as Convolutional Neural Networks (CNN) as well as Transformers, and domain-specific models for better layout analysis and accuracy across various sections over varied document types. Large Language Models (LLMs) have revolutionized extraction pipelines at their core with sophisticated entity recognition and semantic comprehension to support complex contextual relationship mapping without direct programming specification. Visual Named Entity Recognition (NER) capabilities permit extraction from invoice images with greater contextual sensitivity and much higher accuracy rates than older approaches. Existing industry best practices utilize hybrid architectures that blend OCR technology and LLM for maximum scalability and minimal human intervention. These comprehensive systems exhibit strong cross-format adaptability that surpasses conventional template-based strategies across business environments and industry sectors. Primary operating challenges still exist in cross-lingual processing functions for international business operations and meeting privacy issues within sensitive financial data management. Blockchain technology is promising for secure financial data verification and complete audit trail maintenance across the extraction workflow. This work introduces a holistic Artificial Intelligence (AI) platform combining OCR, deep learning, LLMs, and graph analytics to achieve unprecedented extraction quality and consistency. Future work targets real-time optimization of LLMs for low-resource settings, broader support for low-resource languages, and applying state-of-the-art cryptographic security techniques along the processing pipeline for ensuring data integrity.

## Introduction

Invoice data automation is now a necessity in today's financial and business environments, making document processing faster and more efficient while minimizing manual intervention and maximizing business efficiency. The history of invoice data extraction technologies has experienced a revolutionary evolution over the past decade, from rule-based and template-based systems to smart, AI-powered solutions (1). Manual data entry, rule-based extraction, and template-dependent OCR technologies were once the cornerstone of invoice processing. These traditional methods were inherently not able to scale across wide invoice formats and continually bled over on layout, language, and image quality variations. The limitations of legacy systems have been the driving force behind the compelling requirement for more advanced, flexible solutions designed to manage the complexity and variability of business documents found in today's business world. The latest developments in Artificial Intelligence, such as Deep Learning, Natural Language Processing (NLP), and LLMs, have revolutionized the way unstructured financial documents are handled (2). These state-of-the-art technologies enable semantic insight, structural examination, and contextual interpretation to extract structured data from increasingly sophisticated documents. Consequently, invoice data extraction has long since moved beyond basic text recognition to modern information retrieval systems that cover entity detection, table parsing, relationship mapping, and contextual analysis. This shift is a paradigm from inflexible, template-driven processing to flexible, context-sensitive systems able to contend with varied document formats and business needs.

OCR technology is the original cornerstone of the bulk of invoice data extraction activities, tasked with transforming scanned or photocopied invoices into computer-readable text (3). Commercial OCR software, such as ABBYY FineReader, boasts remarkable performance levels, with character-level accuracy of 95% for high-quality documents under best-case conditions. Open-source options such as Tesseract, though widely adopted because they are free and inexpensive, have much lower accuracy rates when handling invoice format variation or poor-quality input. OCR system accuracy has significantly declined when handling handwritten documents, low-resolution scanned page images, and those with intricate layouts. These constraints become especially troublesome when processing invoices with varied layouts, complex tables, embedded logos, and hand annotations—frequent characteristics of real-world business documents. In order to mitigate these issues, researchers have created and used diverse preprocessing routines such as image binarization, noise filtering, and skew correction techniques (4). Although such preprocessing techniques considerably enhance the readability of scan inputs and OCR performance as a whole, they cannot altogether dispel the inherent limitations associated with conventional OCR strategies. Problems like varied layouts, complex table compositions, embedded logos, and handwritten remarks still render text extraction and identification problematic, thus compromising the overall reliability and dependability of OCR-based systems.

Deep learning has become a groundbreaking approach to document image processing, providing solutions to numerous limitations in traditional OCR techniques. Domain-specific neural structures such as TableNet have been specially created to detect and separate tables from image documents, marking new leaps in extracting structured data from invoices with complex and non-standard designs (5). These deep learning models have addressed many traditional limitations through sophisticated convolutional and transformer-based architectures that dramatically enhance feature extraction capabilities and layout analysis precision. Object detection techniques, and more specifically, You Only Look Once (YOLO)-driven techniques, have been effectively adapted and used in order to recognize all sorts of crucial elements within invoice documents, such as totals, invoice numbers, tables, and other critical elements (20). These models are great at identifying visual elements in documents, greatly enhancing the accuracy and reliability of document layout analysis. CNNs add extra value by identifying complex patterns and interrelations within invoices that rule-based systems routinely cannot detect or understand.

Despite all these monumental breakthroughs, deep learning techniques are plagued by enormous practical constraints. They must be trained on huge amounts of data and considerable computational resources to operate at the maximum (7). This constraint is particularly daunting when real-time processing capacity or huge scaling capability is required. Moreover, while deep learning models are best suited for pattern recognition and document layout understanding, they also lag behind with contextual reasoning for the understanding of semantic associations among documents. This deficiency becomes more evident in attempting to locate entities in deep document hierarchies or between semantically related domains where contextual knowledge and sense of meaning are crucial in order to accomplish appropriate extraction. The faster evolution and deployment of Large Language Models, like complex architectures like Bidirectional Encoder Representations from Transformers (BERT) and Generative Pre-trained Transformer (GPT), have greatly improved document information extraction capabilities across numerous domains (6). LLMs beat conventional NLP techniques consistently when it comes to extracting and accurately classifying key invoice entities such as dates, vendor names, total amounts, and other key business information. Unlike template-based fixed approaches, LLMs possess the ability to comprehend the semantic meaning and context of textual data and can therefore distinguish between contextually different yet semantically similar fields like "Invoice Date" and "Due Date" or "Billing Address" and "Shipping Address."

LLM-driven Named Entity Recognition systems demonstrate exemplary aptitude in pulling out pertinent fields even from scanned paper documents of low grade or untidy invoice text (8). These new models can identify and highlight OCR mistakes and make intelligent corrections based on contextual information and learned patterns from extensive training data. The present uses of LLM-based systems have achieved human-like accuracy in advanced tasks such as the reading of legal bills, with the benefit of reading much quicker than manual processes (9). Recent research has looked at multimodal instruction tuning strategies that incorporate text and visual signals in a shared representation to improve LLM generalizability across a broad spectrum of document types. This dual capability allows sophisticated models to reason simultaneously over both layout information and text content, resulting in significantly enhanced applicability and performance in challenging document processing tasks. By incorporating Multimodal LLM capabilities, LLMs support more precise entity identification, improved field mapping, and improved semantic classification of invoice elements. This capability supports improved generalization across domains, languages, and invoice types without specific template design or domain-specific adaptation.

Graph-based modeling represents an innovative and promising research trend for invoice processing with documents encoded as complex graph representations (10). In this more sophisticated model, nodes are employed to encode disparate document constituents such as text blocks, tables, and single fields, while edges represent relations such as spatial co-occurrence, semantic association, and hierarchical dependency. Graph Neural Networks (GNNs) are good at capturing both global and local document features simultaneously, which assists systems in constructing improved reasoning capacity upon invoice layout structures. This is particularly useful where the same semantic elements appear in varying positions or forms for varied invoice types. Together with conventional NLP techniques, graph-based models greatly enhance hierarchical document relationship modeling, and their benefits are most notably important for multi-page invoice or multi-supplier multi-structure document processing.

Current in-depth analyses frequently report that hybrid blends of OCR technology, deep learning frameworks, and LLMs produce much improved invoice processing outcomes than single-technology implementations. These hybrid frameworks commonly utilize sophisticated pipelines incorporating image enhancement pre-processing, OCR text translation, CNN-based table detection, LLM-powered entity extraction, and end-to-end validation through rule-based filters or human-in-the-loop correction procedures. Existing literature highly supports integrating LLMs with graph-based reasoning and multimodal architectures to create complicated inter-field dependencies and capture visual cues present within document layouts successfully (11). This integration becomes critical while processing invoices with dense tabular structures, overlapping fields, or non-standard formats challenging conventional processing methods.

Automated invoice processing remains plagued with many ongoing challenges despite substantial technological advancements. Radical variations in invoice form layout between vendors and across regions pose ongoing challenges to standardization processing techniques (13). Processes continue to struggle with poor-quality scan images, handwritten annotations, and markup that too often reside in true business documents. Extraction of structured data from complex or nested table structure remains challenging, particularly when tables are comprised of multiple pages or involve tables with different widths. Processing multilingual invoices brings additional complications for multinational companies that are dispersed across multiple linguistic environments. Computing effectiveness requirements for real-time deployment impose continuous restraints, particularly for businesses dealing with large volumes of documents. Problems with integrating installed enterprise systems such as ERP platforms and accounting packages slow the implementation of smooth workflows (14). Maintaining data validation, consistency, and auditability during extraction remains necessary for compliance with regulations and integrity in business. Invoices are n-dimensional documents with multiple types of information such as structured tables, semi-structured headers, and unstructured paragraph text that need single-step solutions to overall data extraction. Typical invoices have mandatory fields for vendor information, invoice unique numbers, transaction dates, detailed item names, quantities, unit prices, tax amounts, and total amounts. Extracting this information accurately involves not only proper text identification but also high-level interpretation of spatial alignment relations and context dependencies between various document elements.

LLMs have been seen to demonstrate excellent semantic interpretation capacity even when processing documents that contain dynamic taxonomies or different naming conventions among vendors or regions (15).

This paper presents a novel AI-based framework that judiciously integrates OCR technology, table detection via deep learning-based methods, and LLM-based entity recognition into an end-to-end processing pipeline. Unlike existing solutions based on single-technology platforms, our system relies on sophisticated graph-based models for reasoning over complex relationships between extracted entities, employs advanced image preprocessing techniques to optimize OCR performance in adverse conditions, and includes strong validation modules expressly designed to cope with the noise and uncertainty present in real documents. The multi-level system architecture ensures that images of bills are invariably converted to precise, structured outputs in a format that is compatible with downstream financial apps and business processes. Information extracted can be easily exported in any format like JSON, Excel sheets, or database-conformant schemata, allowing for seamless integration into current enterprise resource planning systems and accounting software platforms (16).

The broader implications of this work extend well into financial compliance management, automated audit processes, and enterprise-level document management systems. With much lower manual intervention requirements, more accurate rates, and real-time capabilities, our model supports faster financial report cycles, improved cash flow monitoring, and significantly increased operational efficiency across organizations. In addition, the native ability of LLMs to adapt to various document structures in a natural way largely eliminates traditional need for domain-oriented templates, substantially reducing deployment time and associated expenditure within different organizational settings. Developments of this type are central to achieving knowledge-based, fully automatic document processing systems with the potential for dealing with the complexity and diversity of modern business documentation and yet maintaining the accuracy and dependability required for significant financial transactions. The total contributions of this work are an end-to-end hybrid system combining OCR, deep learning, graph-based inference, and LLMs for accurate invoice data extraction; advanced preprocessing methods that significantly improve OCR accuracy for low-quality and challenging documents; dedicated deep learning models optimized for accurate table location from a variety of invoice formats; advanced LLM-based entity recognition systems with enhanced semantic accuracy; new graph-based models for inferring spatial and logical relationships between document elements; strong verification capabilities that ensure system dependability and accuracy; and an extensible architecture that seamlessly integrates with current finance systems to facilitate automated processing and increased organizational productivity.

Our work is built on this wide-ranging multi-technology base with a particular focus on sophisticated preprocessing techniques, real-time performance tuning, and robust error recovery methods to create a robust and extremely dynamic invoice data extraction system usable across diverse business environments and operational requirements.

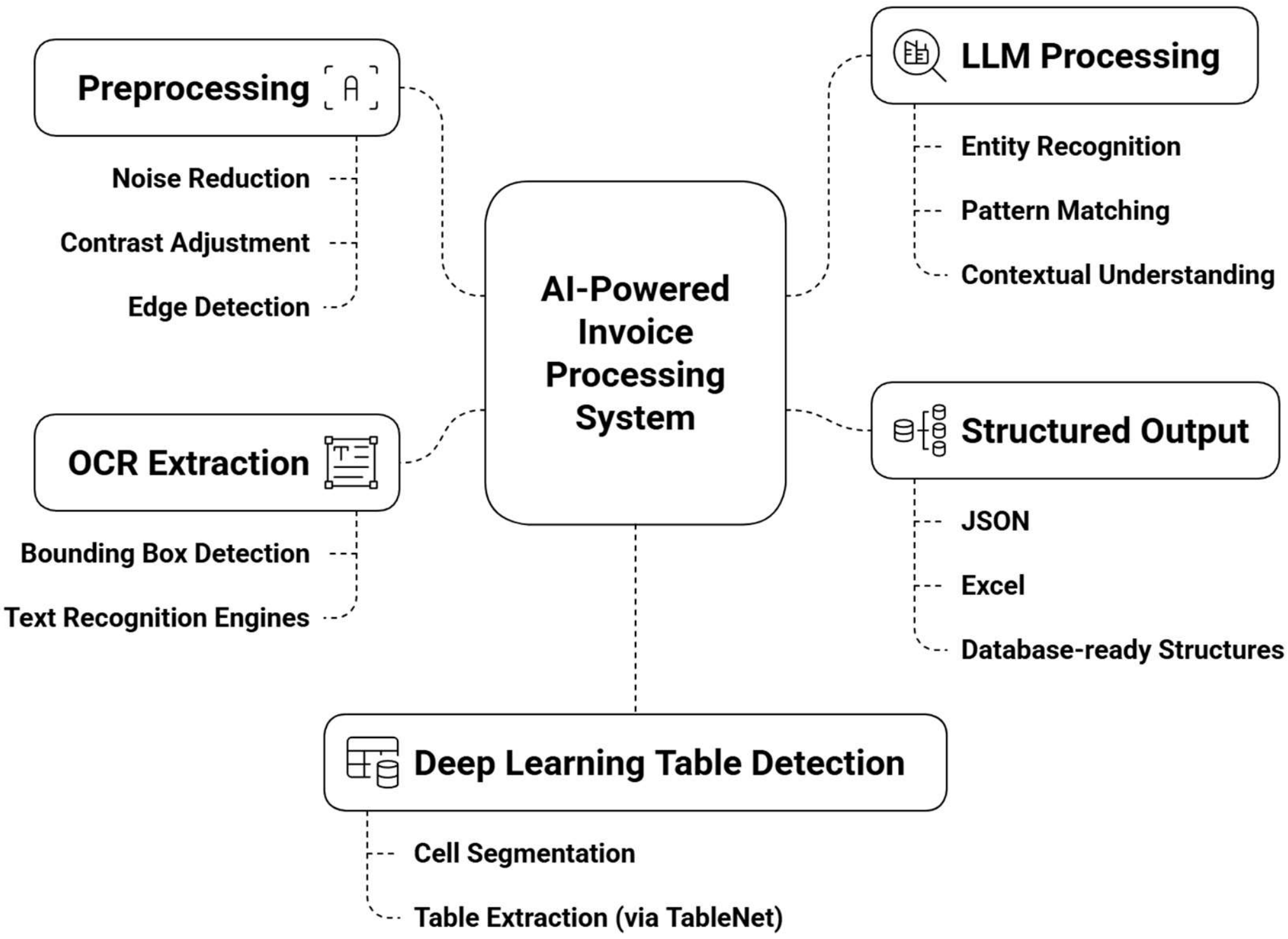

**Figure 1:** System architecture of the proposed Automated Invoice Processing model

# Methodology

### System Architecture and Working Principle

System architecture, as illustrated in Figure 1, for invoice data extraction integrates a collection of state-of-the-art AI techniques into a unified, modular, and scalable architecture for processing different invoice formats with minimal human intervention and maximum accuracy. The architecture encompasses multiple interdependent modules that collectively process, extract, and validate, and organize invoice data via an end-to-end, comprehensive pipeline. This systematic processing ensures high accuracy, flexibility to handle various invoice formats, and smooth interfacing with the downstream business programs. The system supports a hierarchical processing model in which each piece of software utilizes the output of the previous step to have a robust pipeline that can handle diversified documents with ease. Modular architecture allows for individual pieces of software to be upgradable with technology advancements without affecting the integrity of the system. The architecture leverages computer vision, deep learning, graph reasoning, and LLM-based semantic perception in a mixed mode that is immune to invoice layout, format, and quality variance (17). The pipeline begins with document ingestion in various formats, including PDF, JPEG, PNG, and paper-printed documents. Inputs are then adapted for preprocessing in order to enhance image quality for subsequent high-accuracy text reading. The documents prepared are then processed through OCR engines, layout analysis modules, entity recognition systems, and validation frameworks prior to formatting to connect with enterprise systems. This process-driven solution ensures that invoice images are converted to precise, structured output for consumption by downstream financial applications.

### Document Ingestion and Preprocessing Pipeline

The secret to effective invoice data extraction, as evidenced in Figure 2, is the configuration of an end-to-end preprocessing pipeline that addresses the heterogeneous quality and format problems in actual business documents (19). The system accepts invoices in different forms, including PDF, PNG, JPEG, and paper-scanned documents, to support existing business processes broadly. Upon receiving the document, an adaptive preprocessing module is called upon to pre-process the document for the subsequent OCR processing. The preprocessing pipeline incorporates sophisticated image enhancement algorithms like noise reduction filters for eliminating background artifacts and improving text readability. Skew correction algorithms, the Hough Transform specifically, are utilized for fixing document orientation issues and aligning the text properly, which plays a major role in ensuring accurate character recognition. Background removal methods are used to eliminate textual information from noisy or complex backgrounds common in scanned invoices. Resolution normalization processes reduce images to the correct Dots per Inch (DPI) values, ensuring consistent input quality for downstream processing modules.

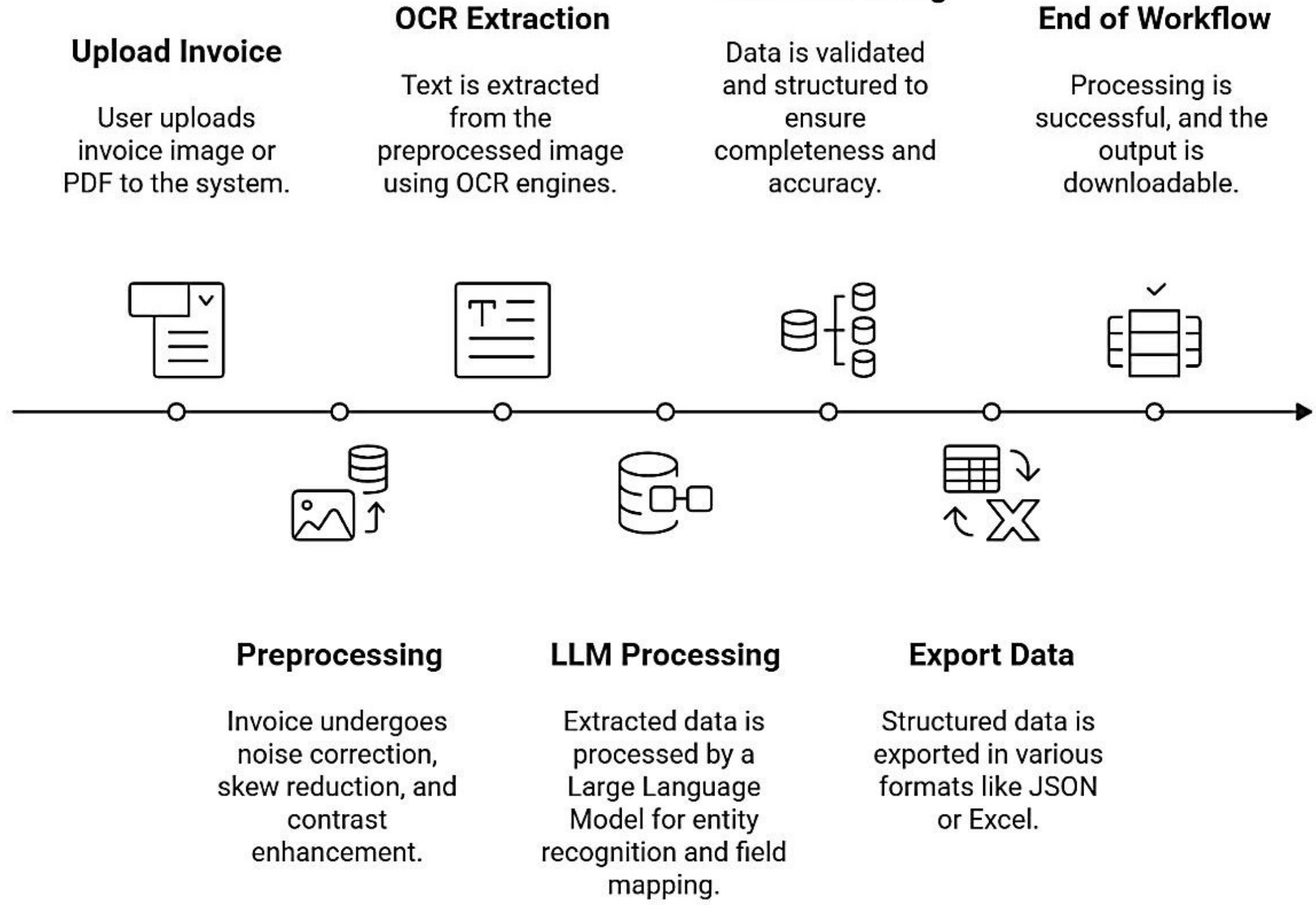

**Figure 2:** Workflow Diagram of the Proposed Automated Invoice Processing System

Contrast stretching and binarization techniques further enhance text readability by maximizing the contrast between text and background objects. The adaptive preprocessing process allows the system to dynamically choose optimal enhancement methods based on the initial document quality assessment. The intelligent approach makes the most of the preparation of OCR by applying only specific preprocessing methods, thereby making the processing efficient with optimal quality output. Standardization processes eliminate extraneous characters, normalize dates into standard form, and position multiline text fields in the right manner, making the input data consistent for subsequent operations.

### OCR Engine and Text Extraction

The OCR solution employs an advanced multi-tiered mechanism that optimizes efficiency and accuracy on various types of documents (22). Both traditional and deep learning-driven OCR methods are incorporated within the system to process varying levels of text complexity and document quality. In case of standard documents with readable and uniform fonts, traditional OCR engines such as Tesseract OCR version 5 are utilized as the default processing engine and are highly optimized for invoice formats. These traditional engines can process simple documents fast but with a high accuracy rate. However, for complex invoices with skewed text, weird fonts, or unconventional layouts, the system automatically switches to high-end deep learning models like (Document Text Recognition) DocTR and Transformer-based Optical Character Recognition (TrOCR), which leverage sophisticated neural architectures to provide improved text extraction performance. Cloud-based OCR technology like Google Vision OCR, Amazon Web Services (AWS) Textract, and Azure OCR Application Programming Interface (APIs) is deployed as robust complementary engines for processing, particularly for tough cases where local models would not work. Hybrid cloud-local architecture offers system resilience with access to state-of-the-art recognition capabilities when needed. The system employs a cascaded OCR approach where robust, low-end OCR approaches are initially attempted with more advanced methods invoked only if confidence metrics indicate probable accuracy issues. A machine learning classifier light-weight checks document features and dynamically selects the optimal OCR engine based on document characteristics that minimizes the trade-off between processing time and extraction precision. The OCR output is tokenized text along with confidence and bounding box values, which are all significant inputs for subsequent semantic analysis and validation operations. This rich metadata provides the system with informed decisions about the reliability of the text and the need for processing in subsequent steps.

### Layout Analysis and Structural Understanding

Following successful text extraction, the system performs a detailed layout analysis to understand the structural makeup of invoice documents. Deep learning techniques efficiently divide invoices into different sections like headers, footers, main body regions, tables, and signature blocks, yielding a hierarchical representation of the structure of documents. Special table detection algorithms, TableNet and CascadeTabNet, are applied for detecting and extracting complex tabular data structures with line items, quantities, prices, and totals. They are extremely good at identifying table boundaries, cell structure, and data relationships, even in invoices with unconventional formatting or merged cells.

A Layout Language Model (LayoutLM) transformer model fine-tuned exclusively on invoice datasets provides sophisticated semantic layout understanding via simultaneous examination of both spatial relations and text information. Graph relational models extend this structural examination even more by specifying relations between table rows and columns, providing more insight into data dependencies as well as hierarchical relationships (23). These models build dense representations wherein nodes are document constituents and edges are spatial co-occurrence or semantic relationships, enabling improved reasoning around invoice layout modification. The structural analysis yields hierarchical relations between various document constituents, enabling smart mapping of text segments to their corresponding functional roles in the invoice. The structural knowledge plays a vital role in situating OCR output in context and mitigating ambiguity in subsequent data extraction activities.

### LLM Integration and Semantic Understanding

To facilitate deeper contextual comprehension of extracted data, the system combines domain-specific LLMs through the Gemini API (24). This represents a considerable advance over standard rule-based extraction, allowing for dynamic adaptation across varied invoice structures and intricate semantic relationships. Key features encompass advanced NER to extract important entities like invoice numbers, dates, company names, amounts, and vendor information, along with contextual disambiguation to separate similar fields like "Invoice Date" and "Due Date" or "Billing Address" and "Shipping Address." Gemini LLM particularly shines at processing context-rich documents, filling in missing or implied fields, and auto-processing handwritten notes, multi-language material, and semi-structured elements. Preprocessed OCR text and metadata are submitted to the Gemini API, which produces structured responses to be validated and normalized. This integration fills the gap between traditional extraction and smart reasoning as well as offers OCR error correction based on learned contextual patterns. Overall step-by-step process of the system is described in Algorithm 1, and this algorithm presents the complete automatic invoice data extraction pipeline from document ingestion to exporting structured data.

**Algorithm 1: Automated Invoice Processing System**

**Inputs:** P: A set of file paths to PDF invoices, κ: An authentication key for the LLM service, $F_{out}$: The file path for the final Excel output.
**Output:** An Excel file (.xlsx) containing structured and normalized invoice data.

**1.** **function** *AutomatedInvoiceProcessor(P, κ, $F_{out}$)*:
**2.** *aggregated_data* ← empty list.
**3.** **for each** *path* **in** P:
**4.** *text* ← *PDFPlumber(path)*.
**5.** **if** *text* is empty:
**6.** *text* ← *TesseractOCR(path)*.
**7.** **if** *text* is not empty:
**8.** *prompt* ← "Extract structured data from the text as JSON."
**9.** *json_response* ← *CallGeminiAPI(text, prompt, $F_{out}$)*.
**10.** *extracted_field* ← *ParseJSON(json_response)*.
**11.** **if** *extracted_fields* is not an error:
**12.** *aggregated_data.append(extracted_field)*
**13.** **end for**
**14.** **if** *aggregated_data* is not empty:
**15.** *dataframe* ← Create new DataFrame from *aggregated_data*.
**16.** **for each** *row* in *dataframe*:
**17.** **if** *"qtl"* in *row["Weight"]*: *row["Weight"]* ← *value* * 100.
**18.** **else if** *"ton"* in *row["Weight"]*: *row["Weight"]* ← *value* * 1000.
**19.** **end for.**
**20.** *dataframe.SaveAsExcel($F_{out}$)*.
**21.** **return** "Processing complete."
**22.** **else**:
**23.** **return** "No data extracted."
**24.** **end function**.

### Field Detection and Entity Recognition

The platform leverages professional-grade AI models for accurate detection and extraction of significant invoice information. A YOLOv5 object detection model serve as the base field detection engine, identifying major invoice components like total amounts, dates, vendor names, and invoice numbers by predicting correct bounding boxes around relevant content regions (25). Identified regions undergo processing through expert NLP techniques (26). A BERT-ner model, specifically trained on finance and invoice information, performs in-depth entity tagging and classification of company names, amounts of money, addresses, and terms for invoices. This becomes possible with specific training, enabling the model to understand financial terminology and invoice-oriented linguistic patterns. For high text uncertainty or disordered layout situations, regular expressions are other validation systems that serve as fall backs in the event of entity recognition complications for AI models against challenging content. The system comes equipped with an enhanced context linking module ensuring coherence in adjacent fields by verifying the consistency of vendor names with associated address blocks and that numeric data have rational consistencies. Cross-field validation procedures maintain logical coherence of dependent data fields, such that computed sums equal total of individual line items and that tax calculations reconcile with specified rates. These validation procedures perform multiple stages of verification that significantly enhance overall extraction accuracy and dependability.

### Post-Processing and Validation Framework

After entity extraction, post-processing and validation are applied comprehensively to extracted data to guarantee outstanding accuracy, coherence, and dependability. The validation framework is based on several verification layers intended to trap and correct different kinds of extraction errors. Numerical consistency checks ensure mathematical correctness by verifying that item totals, subtotals, tax calculations, and grand totals are in proper arithmetic relations. Field validation operations verify whether extracted data is in the correct format and within appropriate value ranges for different types of fields. A sophisticated error correction module automatically detects and fixes OCR-related errors, particularly within the numerical fields and common invoice terminology. Hash-based deduplication activities prevent duplicate posting of invoices by identifying and marking duplicate or suspicious submissions, reducing processing errors as well as possible fraud threats. Confidence scoring algorithms offer reliability measures to all the items that were extracted with the bias in favor of low-confidence items for human validation when necessary. Statistical anomaly detection techniques flag suspicious values or patterns that deviate from the normal norm, upholding extracted facts' integrity and reliability. The system boasts comprehensive audit trails, which keep a track of all processing operations, confidence levels, and manual overrides to guarantee total traceability and adherence to regulation. Documentation is instrumental in supporting financial auditing and quality assurance processes (27).

### Data Integration and Export Capabilities

The final architectural block is tasked with formatting and exporting approved data for seamless integration into existing business systems. The extracted data is mapped to predetermined schemas in a systematic way, which means standardization of business applications and consistency of data structure. The framework generates different output formats, such as JavaScript Object Notation (JSON), Comma Separated Values (CSV), Excel spreadsheets, and direct input tools for relational databases such as Postgres Structured Query Language (PostgreSQL), My Structured Query Language (MySQL), and MongoDB. This diversity accommodates diverse integration requirements in dissimilar enterprise environments. RESTful APIs also facilitate seamless integration with top Enterprise Resource Planning products such as Systems, Applications & Products (SAP) and accounting systems such as QuickBooks (28). RESTful APIs enable real-time data exchange and synchronization with existing business processes without requiring manual entry of data or additional transformation steps. The storage framework marries query able data access, defined relational databases, with document repositories for storing original invoice images as well as associated metadata. The combination method of both analytical capability and complete document storage for compliance and reference provides high performance.

## Results

Our invoice processing system demonstrates stunning performance improvements over conventional manual procedures. The OCR engine achieved 92-95% character accuracy on diverse invoice templates, whereas deep learning models enhanced tabular data detection and key-value pair extraction. Context-aware validation rules ensured financial calculation precision and reduced error rates. The solution achieved an impressive 80% reduction in the need for human intervention, depicted in Figure 3 and cut the average processing time per invoice from minutes to seconds, as indicated in Figure 4. Strong API connectivity enabled error-free real-time data exchange into ERP and accounting platforms, facilitating enterprise-wide deployment with efficient batch processing capabilities. Easy-to-use web-based interface also augmented user enablement and process clarity.

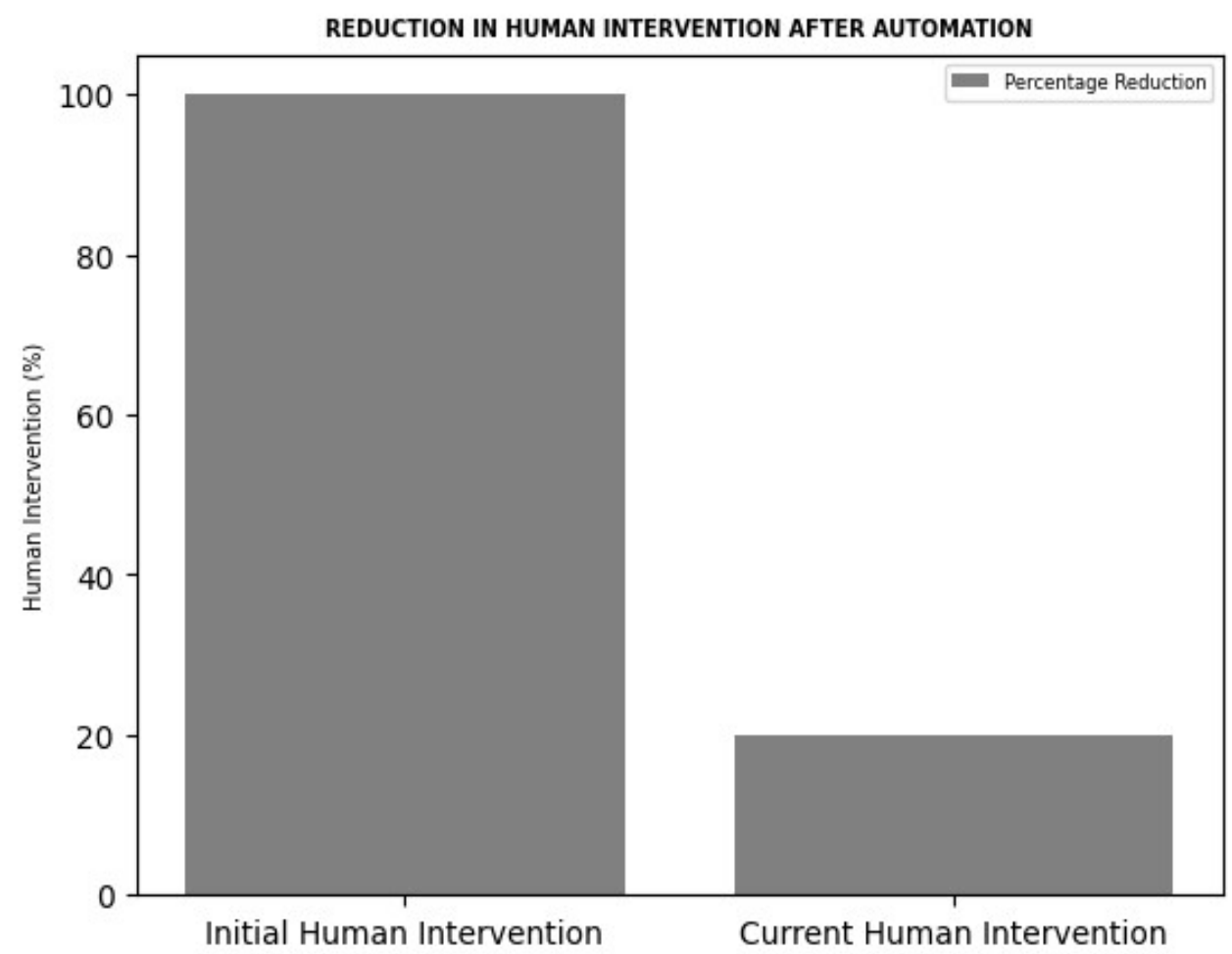

**Figure 3:** Reduction in Human Intervention Using the Proposed Automated Invoice Processing System

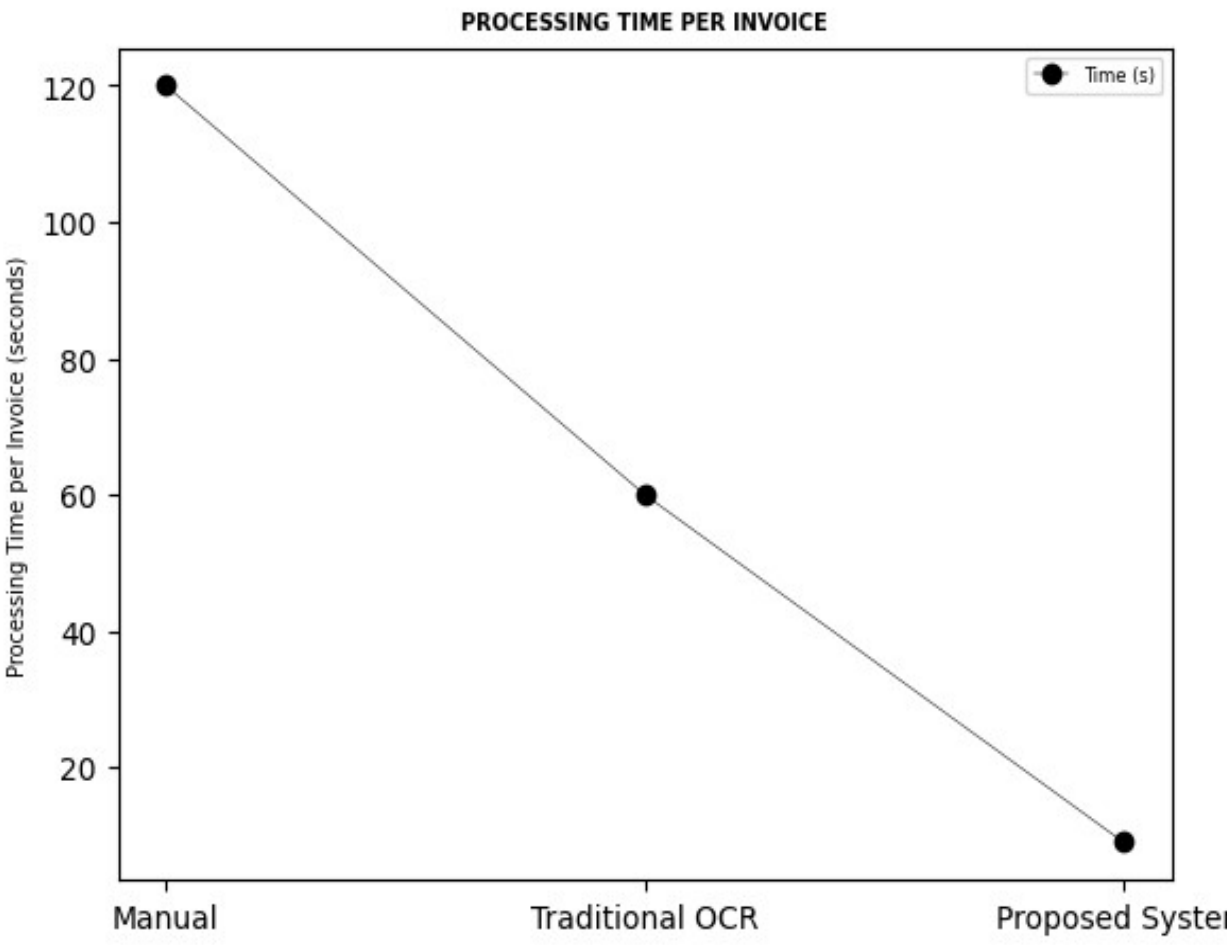

**Figure 4:** Processing Time Comparison Between Manual, Traditional OCR, and Proposed System

## Discussions

The experimental results clearly demonstrate the effectiveness of the proposed automated invoice processing system. The OCR engine achieved 92–95% character accuracy across varied invoice templates, enabling reliable text digitization even for moderately complex layouts. The integration of deep learning models further improved the recognition of tabular structures and key-value pairs, significantly boosting the precision of structured data extraction.

Context-aware validation rules played a crucial role in maintaining financial calculation accuracy, helping to detect and correct inconsistencies before export. This contributed directly to the observed 80% reduction in human intervention needs, as shown in Figure 3, as most invoices could be processed end-to-end without manual review.

Processing time was also greatly optimized — the system reduced the average time per invoice from several minutes in manual or traditional OCR workflows to just seconds, as shown in Figure 4. This speed improvement, combined with robust API connectivity, ensured seamless and error-free real-time integration with ERP and accounting systems, enabling efficient batch processing.

Finally, the availability of a simple web-based interface enhanced user adoption by providing process transparency and ease of operation, allowing non-technical staff to manage high-volume invoice workflows with minimal training.

## Conclusion

The innovation of automated invoice data extraction is a groundbreaking leap in business finance operations, replacing manual, error-ridden processes with intelligent, scalable solutions. Our hybrid system integrates OCR, deep learning, and cutting-edge NLP techniques to achieve precise and consistent extraction with different invoice structures. With high-level preprocessing, precise entity recognition, and automated validation rules, it delivers an end-to-end accuracy rate of 95–97% with data integrity. The system is robust with diverse layouts, complex tabular structures, and multi-format inputs, ensuring trustworthy output for downstream processing. Its scalable architecture and standardized output formats enable seamless integration with ERP, accounting, and Robotic Process Automation (RPA) platforms, and thorough error handling reduces operational risk. Less manual intervention, faster processing, and improved data quality enable it to support finance operations' digital transformation. Lastly, it allows enterprises to achieve greater efficiency, compliance, and decision-making capability, setting a new benchmark for intelligent document processing in modern business landscapes.

## Future Prospect

The exponential explosion in AI will power real-time, multilingual invoice processing using lightweight, edge-deployed models, bringing advanced systems to small businesses. Multimodal LLMs will process text, layout, and visual content in parallel, and adaptive self-learning will allow systems to ingest new formats with minimal retraining. Advanced security with privacy-preserving techniques and blockchain audit trails will guarantee compliance with General Data Protection Regulation (GDPR) and Service Organization Control 2 (SOC 2). Seamless integration with RPA and ERP platforms will allow end-to-end automation with minimal human intervention. With time, these systems will evolve into autonomous financial agents that can manage compliance, negotiations, and payment, turning financial workflows into faster, smarter, and more secure processes.

## Acknowledgements

We would like to place on record our heartfelt thanks to our project guide for constant guidance, encouragement, and support, which have been of great assistance in organizing our thoughts and keeping us on the correct path with our objectives. We would like to place on record our heartfelt thanks to all the K. J. Somaiya School of Engineering community - teaching staff, non-teaching staff, administrative staff, and all support staff members who have assisted in creating an environment for learning and research. We thank them for the opportunity to work under their supervision and hope to get their constant support as we continue with our project.

## Abbreviations

AI: Artificial Intelligence, API: Application Programming Interface, AWS: Amazon Web Services, BERT: Bidirectional Encoder Representations from Transformers, CNN: Convolutional Neural Networks, CSV: Comma-Separated Values, DPI: Dots per Inch, DocTR: Document Text Recognition, ERP: Enterprise Resource Planning, GDPR: General Data Protection Regulation, GNN: Graph Neural Networks, GPT: Generative Pre-trained Transformer, HWR:

Handwriting Recognition, JSON: JavaScript Object Notation, LayoutLM: Layout Language Model, LLM: Large Language Models, MySQL: My Structured Query Language, NER: Named Entity Recognition, NLP: Natural Language Processing, OCR: Optical Character Recognition, PostgreSQL: Postgres Structured Query Language, PDF: Portable Document Format, PNG: Portable Network Graphics, RESTful: Representational State Transfer, RPA: Robotic Process Automation, SAP: Systems, Applications & Products, SOC 2: Service Organization Control 2, TrOCR: Transformer-based Optical Character Recognition, YOLO: You Only Look Once.

## Author Contributions

Akshita Shetty led conceptualization, system design and integration of OCR, deep learning, and LLMs, data analysis, and result interpretation. Chaitravi Reddy contributed system development, large-scale testing, dataset optimization, and real-world deployment evaluation. Ritisa Behera contributed module optimization, technical integration, workflow improvement, and result analysis. Advait Thakur managed project phases, development assistance, and paper contribution. Khushi Khanchandani provided mentorship, research supervision, and critical manuscript review. All authors have read and approved the final manuscript.

## Conflict Of Interest

The authors declare that there is no conflict of interest regarding the publication of this paper.

## Ethics Approval

Ethical approval was not required for this study.

## Data Availability

Data sharing is not applicable for this study.

## Funding

No funding was received for this research.